# Resource Prediction for Humanoid Robots

Manfred Kröhnert*, Nikolaus Vahrenkamp*, Johny Paul‡, Walter Stechele‡, and Tamim Asfour*

* Karlsruhe Institute of Technology,
Institute for Anthropomatics and Robotics
{Kroehnert, Vahrenkamp, Asfour}@kit.edu
‡ Technical University of Munich,
Institute for Integrated Systems,
{Johny.Paul, Walter.Stechele}@tum.de

*Abstract*— **Humanoid robots are designed to operate in human centered environments where they execute a multitude of challenging tasks, each differing in complexity, resource requirements, and execution time. In such highly dynamic surroundings it is desirable to anticipate upcoming situations in order to predict future resource requirements such as CPU or memory usage. Resource prediction information is essential for detecting upcoming resource bottlenecks or conflicts and can be used enhance resource negotiation processes or to perform speculative resource allocation.**

**In this paper we present a prediction model based on Markov chains for predicting the behavior of the humanoid robot ARMAR-III in human robot interaction scenarios. Robot state information required by the prediction algorithm is gathered through self-monitoring and combined with environmental context information. Adding resource profiles allows generating probability distributions of possible future resource demands. Online learning of model parameters is made possible through disclosure mechanisms provided by the robot framework ArmarX.**

## I. INTRODUCTION AND RELATED WORK

Nowadays, the latest generation of humanoid robots is typically equipped with off-the-shelf PCs. Each PC contains a single or multi-core processors (1 - 8 CPU cores) onto which the different (computation) tasks are statically distributed upon design time [1]–[5]. In the current decade however, the roadmap for semiconductors from the Semiconductor Industry Associations (SIA) foretells the availability of many-core processors with hundreds to thousands of cores. To deal with such huge amounts of resources new methodologies such as resource-aware programming are required. Therefore, the transregional research project Invasive Computing is currently developing a resource-aware architecture [6]. Overall, the idea of Invasive Computing is about writing self-adaptive applications by providing an explicit interface for requesting and releasing resources. The notion of requesting and releasing resources is built into the developed platform ranging from hardware over operating system up to programming language support. As part of the project an enhanced parallel operating system (OctoPOS) capable of using hardware-based resource-aware mechanisms is being developed [7]. OctoPOS contains a software-based agent system for managing resources between applications. The agent system is for example responsible for negotiating about free resources and strategic resource planning [8]. This negotiation prior to allocation might grant less resources than requested if the agent system has to balance requests from several applications. Thus, every application must be designed to operate with less resources than requested.

Recent research has shown that resource-aware algorithms used in robotic systems can guarantee a certain degree of quality of service, such as desired throughput, by adapting to continuously changing amounts of granted resources [9]–[11]. This self-adaption can lead to increased overall system stability and failure tolerance as well as better load balancing.

When a robot needs to adapt its current behavior to changes in its highly dynamic surroundings, such self-organization techniques are key for efficient resource allocation and usage. It is desirable to anticipate upcoming situations to predict future resource demands in advance of executing actions. Predicted resource requirements can be passed on to an invasive runtime system where they are used by the agents as additional hints to consider during resource negotiation. Since predictions range multiple time steps into the future, taking them into account while negotiating on available resources allows the agent system to optimize with respect to a longterm goal such as avoiding upcoming resource bottleneck. If the current resource situation permits, it would even be possible to perform speculative resource allocations for upcoming tasks.

First, this paper describes how the state disclosure mechanisms of the robot framework ArmarX [12] are expanded to gather execution traces of robot programs through self-monitoring. These traces are enriched with environmental information to generate context dependent data sets. Next, a prediction model based on Markov chains is presented which is capable of learning and adapting its parameters from previously performed executions. The developed model is used in a human-robot interaction scenario to predict possible future behaviors of the humanoid robot ARMAR-III [3] based on the current context. Additionally, the calculated prediction is combined with a resource model to provide a preview of potential future resource demands. Future resource demands are expressed as a probability distribution instead of only the best matching prediction. This allows an agent system to take into account multiple diverging resource demands weighted by their probability.

The remaining part of this section describes related work with regards to monitoring concepts, prediction methods, and resource models.

The Task Description Language from Simmons and Apfelbaum contains a *Monitor* task which is executed periodically [13]. Once a previously defined condition is met, an event can be triggered which other tasks can react upon.

Steck et al. introduced a model-driven approach to robot software design [14]. It includes mechanisms for monitoring available resources at runtime. First, resource usage of each robot program is modeled at design time. At runtime the activation/deactivation of the programs as well as the wiring between programs can be changed depending on the monitored resource situation.

A similar approach has been presented by Park et al. [15]. In this system, monitored resource usage patterns are the basis for optimizing the planing of following actions. The described system is capable of extracting the current resource situation through reference counting of active components and generates matching resource constrained plans afterwards.

A hybrid model for execution and self-monitoring is introduced in [16]. In this work, the execution of a robot is monitored with a heuristical model. Obtained datasets are used in a learning phase to train an instance-based model mapping sensor information to robot behaviors. The learned instance-based model is then used to detect the progress of a robot in a navigation task. Possible gaps in the instance-based model are filled by consulting the heuristical model.

Additionally, it is necessary for a robot to detect and adapt to changes in its environment as explained in [17]. In this work, a survey of current execution monitoring methods in robotics was conducted based on the need of a robot to monitor its execution to be prepared for detecting failing executions. Several categories of monitoring approaches are described while an observer-based monitoring is presented to be used most commonly due to its correlation with state machine representations of robot programs.

Machine learning methods for calculating predictions are described in [18]. Listed are algorithms such as Bayesian Learning, Hidden Markov Models (HMM), and Support Vector Machines (SVM).

Learning a model of a navigation task is based on Dynamic Bayesian Networks (DBN) in [19] . The DBN can predict future situations as well as handling objectives such as avoiding failure states.

Markov chains model the behavior of tasks in an image processing pipeline in [20]. Switches in the control flow of the program are most reliably predicted and used to calculate upcoming resource usage patterns.

Prediction algorithms for situation interpretation are vital for cognitive automobiles. Case Based Reasoning (CBR) is one approach selected in [21], and a Hidden Markov Model based method is described in [22].

Prediction models are also an important aspect in software engineering, amongst others in the work of Kounev et al. An enhanced software architecture model capable of answering performance queries at run time is described in [23]. Specific combinations of resource allocation and load-balancing strategies are evaluated based on the execution of such performance queries.

Several resource models are described in literature. A simple model for resources is presented in [15]. Both CPU and memory consumption of all available robot tasks were measured to reuse them during the generation of resource constrained plans.

The Palladio Component Model (PCM) originates from the field of model-driven software development [24]. It contains a resource model for basic performance evaluation of a software architecture during design time. The resource model is restricted to basic types such as processor and communication resources.

A more general resource description can be found in the *UML Profile for Schedulability, Performance, and Time* provided by the Object Management Group (OMG) [25]. This model covers a broader variety of available resources than PCM.

## II. SYSTEM OVERVIEW

The objective of this section is to give a brief overview of the developed system before describing implementation details in the next section. We will discuss monitoring concepts and context knowledge modeling, as well as prediction, online learning and the resource models required for calculating future resource demands. Finally, an evaluation scenario will be presented.

### A. Monitoring

The previous section introduced several papers working with self-monitoring. Most of them are limited to sensor values such as speed or position of a robot.

This work is based on the ArmarX Robot Development Environment. In contrast to other classical self-monitoring approaches, ArmarX emphasizes on state disclosure as essential requirement to implementing robot capabilities and applications [12]. Within ArmarX, a robot program is formally represented by state transition networks (hierarchical statecharts) and state-transitions are triggered by events. This structured representation of a sequence of states is exposed to external tools and provides means for intuitive inspection of the system state of a robot program. Estimation and prediction of the execution state of a robot as well as profiling and monitoring are performed on a high level of abstraction.

As opposed to [16], the current robot state is not predicted based on monitor sensor values. Instead, the current state $s_c \in S$ of the robot used for prediction is determined from the set of all available states $S$ through monitoring of the robot program execution. Since all states are implemented in a generic way, they need to be parameterized before being executed. The associated state parameters are stored in a set $\phi_x \in \Phi$, with $\Phi$ containing already observed state parameter sets. Furthermore, transitions between states are stored together with a transition count.

### B. Context Knowledge

The memory structure of the ArmarX framework provides context knowledge. It contains information about the robot environment on a symbolic level and allows for making queries as well as registering for updates triggered by changes made to the memory. This context knowledge is created by taking a snapshot of specific environmental parameters and storing them in a database for later retrieval. These parameters are stored in a set $\psi_x \in \Psi$ (collection of already observed environmental parameter sets) and linked with stored transitions. Later on, these context parameters are combined with the current execution state of the robot to serve as input of the prediction model.

### C. Prediction Model

Before choosing a viable prediction model, a characterization of the underlying environment model is required. The following environmental properties are assumed in the scope of this paper:

- *dynamic*: the environment can change without interaction of the robot
- *discrete*: statechart as well as environmental object states are discrete
- *stochastic*: external events may occur with a certain probability (e.g. a human entering the scene and interrupting the robot)
- *completely observable*: the state of all objects in the scene is known and can be observed

Since the goal is to make context-aware predictions it is required to take the world state $\omega_i$ into account. It is defined as

$$\omega_i = (\phi_i, \psi_i, s_c), \ \phi_i \in \Phi, \psi_i \in \Psi, s_c \in S$$

All observed world states $\omega_i$ are collected in $\Omega$.

The prediction of the next probable world state $\omega_{t+1}$ reachable from the current world state $\omega_t$ is defined as

$$p(\omega_t) = \omega_{t+1}, \ \ \omega_t, \omega_{t+1} \in \Omega$$

Since the environmental model is discrete and completely observable, the amount of possible world states is finite. Therefore, it is possible to define the prediction model as a first order Markov chain. The probability vector $X \in \mathbb{R}^n$ describes the transition probabilities $\pi_i$ from world state $\omega_i$ to any other world state observed so far. Every $X$ vector contains probabilities of the $n$ observed world state entries and holds the following properties

$$\pi_i \geq 0, \ \ \sum_{i=0}^{n} \pi_i = 1$$

The transition probability from world state $\omega_i$ to $\omega_j$ is defined as

$$p_{ij} := P(X_{t+1} = \omega_j | X_t = \omega_i), \ \omega_i, \omega_j \in \Omega$$

The transition matrix $M$ from any world state to any other world state and is then described as

$$M = \begin{pmatrix} p_{11} & \cdots & p_{1n} \\ \vdots & \ddots & \vdots \\ p_{n1} & \cdots & p_{nn} \end{pmatrix}$$

Since $p_{ij}$ is independent of time parameter $t$ from $X_t$ the model can be described as a homogeneous Markov chain. Therefore, a prediction horizon $h$ can be specified, enabling the calculation of predictions of up to $h$ time steps into the future.

Given a world state $\omega_i$ the prediction is calculated by executing the following steps:

1) Create vector $X_i$ by marking all entries 0. Only the entry identifying $\omega_i$ is marked 1.
2) Construct the transition matrix $M$ from the latest database entries according to

$$p_{ij} = \frac{count(\omega_i \rightarrow \omega_j)}{count(\omega_i \rightarrow x)}, \ x \in \Omega$$

(number of transitions from $\omega_i$ to $\omega_j$ divided by all transitions starting at $\omega_i$).
3) Calculate the new probability distribution for $\omega_j$:

$$X_j = X_i^T * M$$

4) Pick the entry with the highest probability as the next world state and extract the related robot program state.
5) If prediction horizon $h$ is greater than 1 repeat the calculation of $\omega_j$ $h$-times while replacing $X_i$ with $X_j$ in each step.

### D. Online Learning

Initially, the prediction model is empty and all transition probabilities are assumed to be equally distributed (regardless of the world state). Each gap in the prediction model generates a prediction based on the equal distribution upon first encounter. While statechart transitions are encountered during the execution of a robot program, they are stored in a database accompanied by a snapshot of their respective environmental properties. These database updates are key to continuous updates of the probability values in the transition matrix $M$. Additionally, gaps in the model are closed on-the-fly and newly observed situations are taken into account immediately.

### E. Resource Model

A reduced resource model describing both CPU usage (between 0 and 100 %) and memory usage (between 0 and 1000 MB) is used in this work. For each relevant state in the statechart a pair of CPU/memory values was generated based on expectations of the corresponding algorithms. These resource profiles are then associated with calculated predictions to forecast future resource demands. In the future, the described resource models will be replaced with information obtained from profiling the algorithms on the real robot hardware.

### F. Evaluation Scenario

The prediction of actions in a Pick and Place robot program based on ArmarX is evaluated to assess the performance of the presented model. The task of the robot is to pick up an object from a table and place it elsewhere (the sideboard for example). Without interruptions, the task is straightforward:

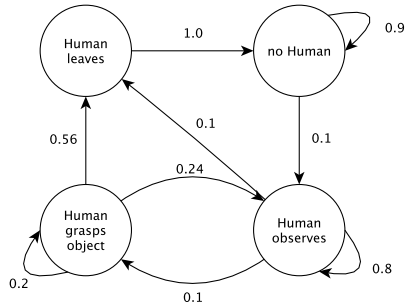

Fig. 1. Transition probabilities of human actions when entering the workspace of the robot.

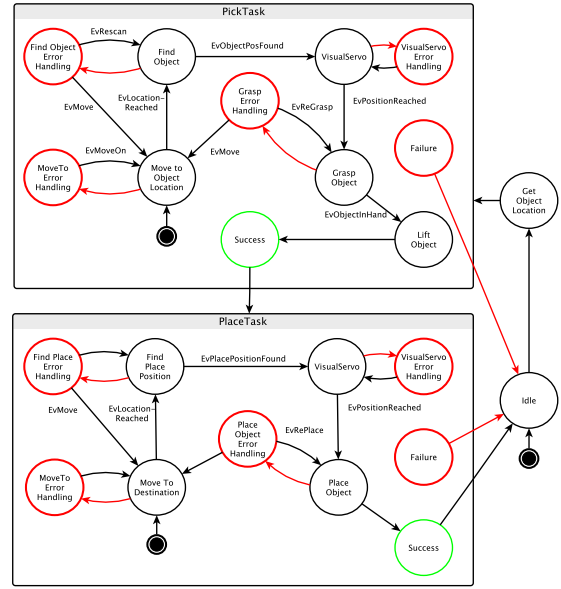

Fig. 2. The complete Pick and Place statechart. Error states and related transitions are colored red while `Success` states are green. Transitions going to the `Failure` states have been omitted from the diagram to focus on the main execution flow.

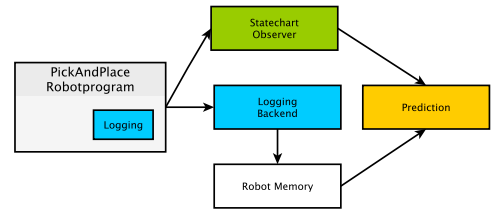

Fig. 3. Overview of the components involved in the prediction progress.

go to table, locate object, reach object, grasp and lift object, go to different location, place object, release grasp, and lift hand. However, it is possible for a human operator to enter the scene during execution and perform one of the following tasks:

- with a high probability the human will enter the room and stay near the entrance
- walk to the table and grasp an object (possibly be the robot's object)
- leave the room again

The probabilities for each of the describe actions are depicted in Fig. 1.

If the human intents to grasp an object on the table, the robot needs to react upon the interruption to prevent damage from the human. This can either result in aborting the current task and starting a dialog with the user to ask for a new task, or in continuing the grasping action if the human grasps an object on the other side of the table. Once the robot encountered a person multiple times performing a similar action, the behavior will be encoded in the saved world states and will influence the prediction system in the future.

The complete statechart including error handling states is depicted in Fig. 2. The robot program starts in the `Idle` state from where it receives a pick and place command. The `PickTask` is entered after the location of the object to be fetched is known. Finishing the `PickTask` with entering the `Success` substate results in a transition to the `PlaceTask`. Some error conditions are handled explicitly with dedicated error states. All other errors during execution will force a transition to a dedicated `Failure` substate while aborting all current actions. To preserve the clarity of the diagram, all transitions ending in `Failure` have been omitted in favor of a better overview of the main execution flow of the statechart.

## III. IMPLEMENTATION

As mentioned in the previous section, both self-monitoring and prediction have been implemented using the ArmarX framework. The implementation comprises three major additions to the framework: Statechart Logging, Statechart Observation, and Prediction. Dependencies between these components are visualized in Fig. 3.

### A. Statechart Logging

All transitions in an ArmarX based statechart are triggered by events. Every parent state receives and dispatches events to manage transitions between its child states.

To support monitoring of transitions within a statechart, a pluggable `StatechartLogger` has been implemented which can be inserted into the event processing loop. This `StatechartLogger` gets called every time a transition is performed while the logger is active. Parameters are used to enable/disable logging, as well as to specify which states are considered important. For example, the execution of the visual servoing state – as opposed to the substates required to implement it – is considered important. Currently, expert knowledge is required to specify the important high-level states of the statechart.

After a transition is detected, it is sent to a logging backend through a publish subscriber mechanism. This mechanism allows for distributed sending of transition information as well as processing the information with multiple logging facilities. Once a state transition is received by a logging backend, the memory is queried for additional environmental parameters. If a matching world state is found in the database the respective transition counter is incremented. Otherwise, the gathered information is stored as a new world state $\omega_x$ in the database.

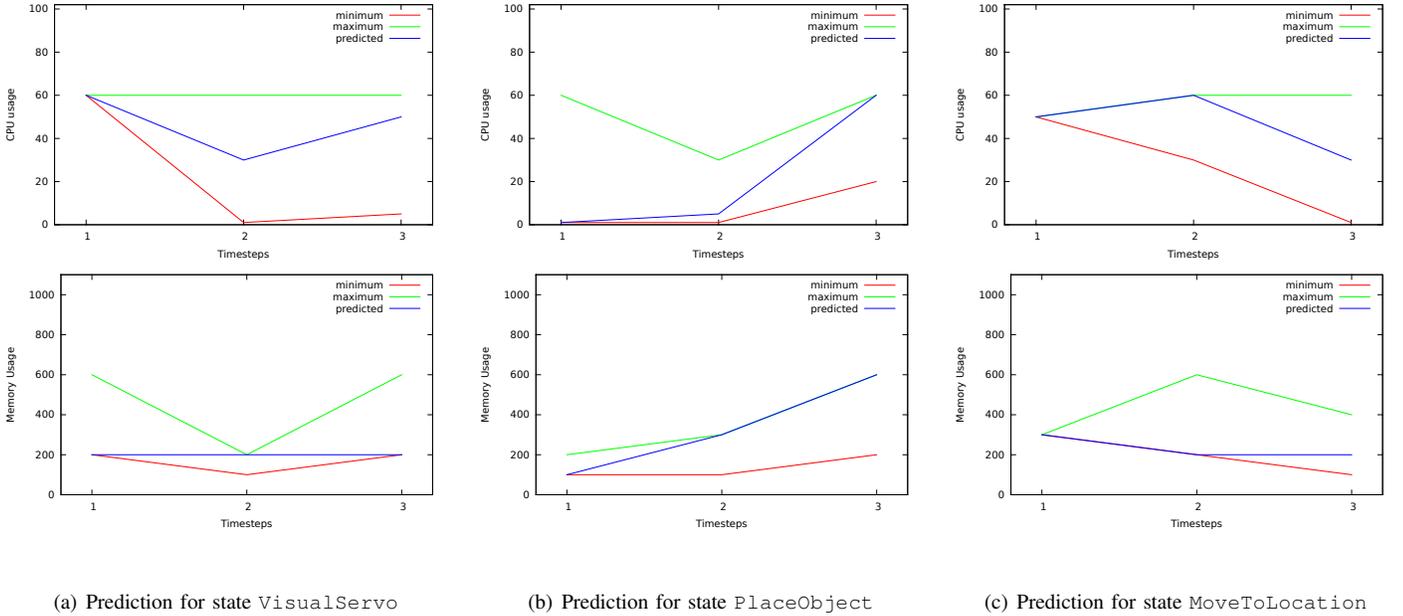

(a) Prediction for state `VisualServo`

(b) Prediction for state `PlaceObject`

(c) Prediction for state `MoveToLocation`

Fig. 4. Resource profile predictions generated during execution of the Pick and Place task. Each subfigure shows the predicted CPU usage in % (top) and memory consumption in MB (bottom) for a different substate. Maximum values are shown in red and minimum values in green. Values from the most probable prediction are colored blue.

### B. Statechart Observation

One recurrent pattern in the ArmarX framework is the `Observer` monitoring any kind of data and providing appropriate data channels. During runtime, a state can add conditions such as $sensorvalue \geq 200$. At the same time, an event is attached to each condition added. On condition fulfillment the associated event is sent back to the statechart which added the respective condition. The `StatechartObserver` (deriving from `Observer`) has been implemented to provide a data channel holding the currently active state. It is then possible to register a `stateChanged` condition on the provided channel which triggers an event every time the active state changes. Similar to the logging mechanism it is possible to select the monitored states individually.

### C. Prediction

The `Prediction` component makes use of the aforementioned condition mechanism. Upon first start, a `stateChanged` condition is registered with the `StatechartObserver`. Every time this condition gets triggered, a snapshot of the current world state is taken and associated with the current execution state of the statechart. Subsequently, the prediction is computed as described in the *System Overview* section by loading all relevant data from the robot memory. The loaded data is then used to construct the transition matrix required to calculate the prediction. Since the Markov chain is homogeneous, it is possible to calculate predictions reaching multiple time steps into the future. The desired prediction horizon $h$ can be specified on startup.

## IV. EVALUATION AND RESULTS

The previously described architecture is evaluated by simulating the execution of the robot in the scenario as described in section II-F. A single Pick and Place task is executed while additional failure events as well as human interruptions are generated at random.

First of all, more than 500 Pick and Place task executions were performed to gather training data for the prediction model in advance of the evaluation. The evaluation was performed by executing the scenario while each transition of the statechart triggered a prediction calculation with a horizon of 3 time steps. For each prediction the associated resource profile was attached and displayed.

The generated resource profiles are shown in Fig. 4. CPU usage (top row images) and memory consumption (bottom row images) are shown for predictions calculated in three selected substates. Red lines indicate lower boundaries, green lines indicate upper boundaries, and blue lines indicate the resource profile of the most probable prediction. To generate these graphs, all transitions with high probabilities are chosen in descending order until their accumulated probability is greater or equal to 0.75. For each time step the minimum and maximum resource values are displayed together with the most probable one.

The resulting figures depict predictions of the robot executing `VisualServo`, `PlaceObject`, and `MoveToLocation` states. In case of `VisualServo` (Fig. 4(a)) the `GraspObject` state was predicted most probable with an overall high CPU usage.

The gap in step 2 is a result of `Failure` requiring less resources than `GraspErrorHandling` which in turn requires less resources than `LiftObject`.

Similarly, the gap in CPU usage from Fig. 4(b) results from the difference of the `Success` state requiring no resources and the `PlaceObject` state which might be reentered due to an interruption.

CPU and memory usage in Fig. 4(c) start of with identical values. According to the current world state only `FindObject` could be predicted as follow up state.

Prediction precision of the presented approach was evaluated in 4 executions and results are presented in TABLE I. The actually executed substate was compared to the most probable predicted substate to determine if the prediction was correct. Human interruptions were constantly triggered in each of the four evaluations. The transition probabilities of the human actions are detailed in Fig. 1. Additional failure events were generated with a probability of 0.3 in two of the evaluations (marked `E`) to simulate failures during the execution of single substates. Predictions were calculated in each combination with two different sets of criterions. On one hand, the State-Match (S-Match) criterion only checks if the predicted state matches the actually executed substate. On the other hand, the World-Match (W-Match) criterion takes environmental parameters into account. Predictions are only considered correct if both predicted/executed substate as well as associated environmental parameters match.

TABLE I
PERCENTAGE OF CORRECT PREDICTIONS OF THE MARKOV BASED METHOD. E GENERATES ADDITIONAL FAILURE EVENTS. S-MATCH DETECTS ONLY SUBSTATE MATCHES. W-MATCH CONSIDERS BOTH SUBSTATE AND ENVIRONMENT PARAMETER MATCHES.

| Parameters | S-Match | W-Match | S-Match + E | W-Match + E |
|---|---|---|---|---|
| Precision [%] | 91.3 | 76.3 | 53.1 | 43.5 |

If additional failure generation is enabled, prediction performance drops since the amount of possible transitions is increased with at least one transition per substate. Overcoming this performance drop requires gathering many more training samples.

In both scenarios S-Match always counts more correct predictions than W-Match. Adding environment parameters decreases the prediction correctness. For example, the robot might perform the same action when a human is present. However, W-Match still distinguishes between world states if the present human differs even if both humans perform the same action. Nevertheless, relying on S-Match is not practical in cases where possibly interfering habits of different humans need to be considered. The preference of a human to choose grasping one object over another is vital if this action might interrupt the robot. To increase the number of correct predictions regarding the W-Match criterion it is necessary to do further research on which environmental parameters are essential for generating correct predictions and how to determine those important parameters.

In this initial implementation, the focus lay on calculating the prediction and not on the speed of the prediction. However, the calculation of the prediction took less than one second. This overhead is negligible compared to the execution times of the states that are being predicted. Since the prediction is calculated on a high level of abstraction the execution times of the states are usually in the range of a couple of seconds.

## V. CONCLUSION

In this paper, a new approach is presented combining self-monitoring with resource models to generate predictions about future resource demands of a robot program. An observer-based monitoring approach was chosen as proposed in [17]. Compared to other work, the monitoring goes beyond sensor values and allows collecting additional high level information.

The ArmarX framework was enhanced to support self-monitoring of internal transitions of robot programs. A Markov chain model was trained with monitored execution traces and environment parameters in order to predict the most probable actions of the robot depending on the current world state. Afterwards, the calculated prediction was combined with previously defined resource profiles to provide an outlook on possible future resource situations. Tests were performed in a simulation of the robot ARMAR-III in a fully observable environment. The robot executed a Pick and Place task while interruptions were constantly caused by a designated human operator model.

It was shown that both environment and transitions of ArmarX based robot programs can be monitored with this approach. Predictions were calculated in order to estimate a probability distribution of future resource actions of the robot based on the collected information. Resource profiles were linked to the calculated prediction to show upcoming workload situations.

The execution of a single task not exceeding the current system limits was analyzed. However, system overload is expected to occur when multiple concurrent tasks are executed. With the presented approach, possible overload situations can then be detected in advance. Combining resource prediction with new paradigms such as Invasive Computing will provide means to influence the resource negotiation process in advance of a resource request and let the agent system deal with preventing system overload. This technique is expected to enable more efficient resource utilization and better load balancing between concurrent tasks.

Future work is going to expand the presented scenario into a more complex one running multiple concurrent tasks. More detailed resource models will be generated by profiling applications executed on the humanoid robot ARMAR-III. Additionally, more enhanced prediction models such as HMMs will be evaluated for dealing with incomplete world knowledge.

## ACKNOWLEDGEMENT

This work was supported by the German Research Foundation (DFG) as part of the Transregional Collaborative Research Centre Invasive Computing (SFB/TRR 89).